\documentclass{article}

\usepackage{PRIMEarxiv}

\usepackage[utf8]{inputenc} 
\usepackage[T1]{fontenc}    
\usepackage{hyperref}       

\hypersetup{colorlinks=true}

\usepackage{url}            
\usepackage{booktabs}       
\usepackage{amsfonts}       
\usepackage{nicefrac}       
\usepackage{microtype}      
\usepackage{lipsum}
\usepackage{fancyhdr}       
\usepackage{graphicx}       
\usepackage{tabularx}
\graphicspath{{media/}}     

\usepackage{ifpdf}
\usepackage{caption}

\usepackage{orcidlink}

\usepackage{caption}
\usepackage{subcaption}

\usepackage{amsmath}

\usepackage{multirow}
\usepackage{stackengine}

\renewcommand{\arraystretch}{1.3} 

\title{Tracking Moose using Aerial Object Detection
}

\author{
  Christopher Indris\upstairs{\affilone}\orcidlink{0009-0003-3327-0218}, ~
  Raiyan Rahman\upstairs{\affilone}\orcidlink{0009-0000-1392-6441}, ~
  Goetz Bramesfeld\upstairs{\affiltwo}\orcidlink{0000-0001-7060-8124}, ~
  Guanghui Wang*\upstairs{\affilone}\orcidlink{0000-0003-3182-104X} \\
  \upstairs{\affilone}Department of Computer Science \\
  \upstairs{\affiltwo}Department of Aerospace Engineering \\
  Toronto Metropolitan University \\
  Toronto, Canada\\
  {*}Correspondence: \texttt{wangcs@torontomu.ca} \\
}

\begin{document}
\maketitle
\begin{abstract}
Aerial wildlife tracking is critical for conservation efforts and relies on detecting small objects on the ground below the aircraft. It presents technical challenges: crewed aircraft are expensive, risky and disruptive; autonomous drones have limited computational capacity for onboard AI systems. Since the objects of interest may appear only a few pixels wide, small object detection is an inherently challenging computer vision subfield compounded by computational efficiency needs. This paper applies a patching augmentation to datasets to study model performance under various settings. A comparative study of three common yet architecturally diverse object detectors is conducted using the data, varying the patching method's hyperparameters against detection accuracy. 
Each model achieved at least 93\% mAP@IoU=0.5 on at least one patching configuration. Statistical analyses provide an in-depth commentary on the effects of various factors. Analysis also shows that faster, simpler models are about as effective as models that require more computational power for this task and perform well given limited patch scales, encouraging UAV deployment. Datasets and models will be made available via \href{https://github.com/chrisindris/Moose}{github.com/chrisindris/Moose}.
\end{abstract}


\begin{figure}[!htbp]
    \centering
    \includegraphics[width=\linewidth]{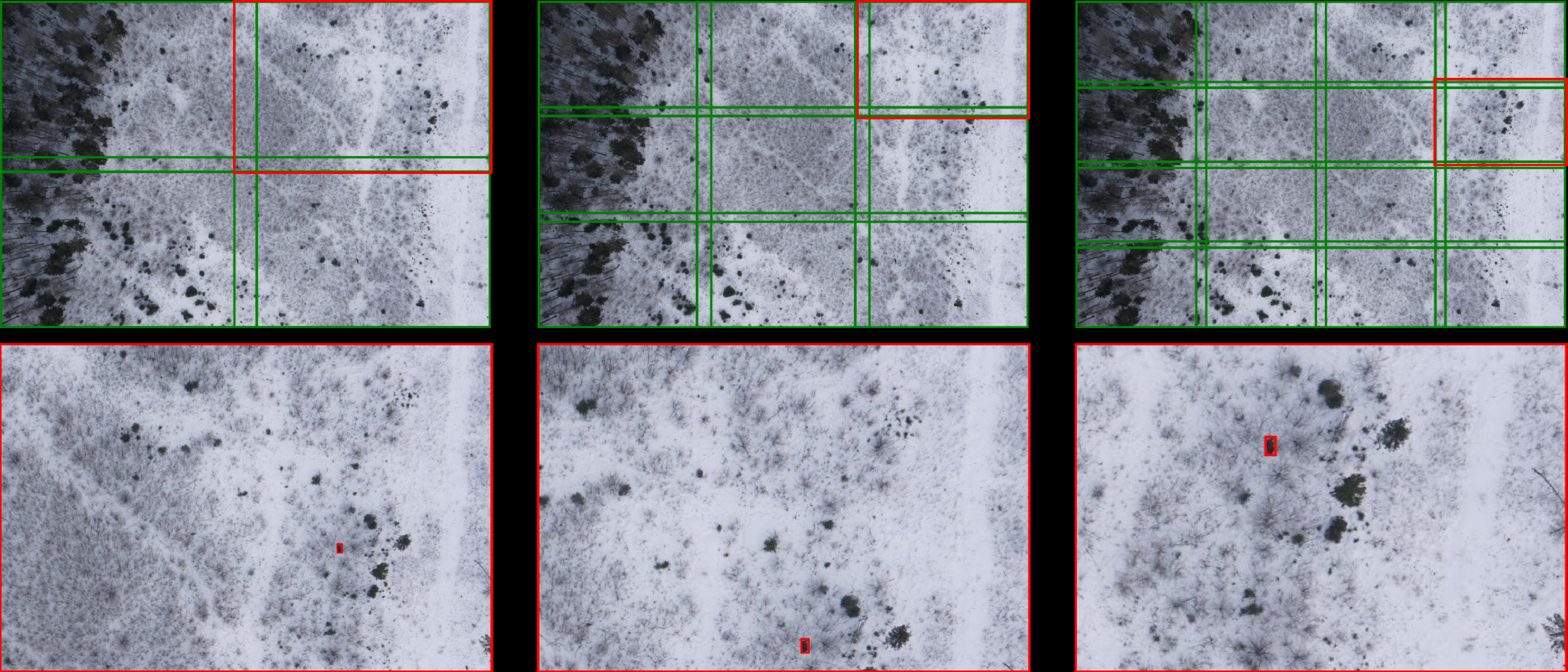}
    \caption{An illustration of the patching method. Left-to-right across the top row shows how an image is divided into large, medium and small patches; note the overlap between them. The bottom row shows each red patch from the image above it, which includes an annotated moose.}
    \label{fig:patchdiagram}
\end{figure}

\section{Introduction}
\label{sec:introduction}

The natural habitats of wildlife have been altered by climate change and other human activities. Included among them are the moose and caribou of Canada's north, two species which are culturally important to many Indigenous peoples and are threatened by changes to their habitat \cite{borish_its_2022, Ford:2009aa}. For informed decision-making regarding government policies and conservation efforts, surveying is essential \cite{johnson_response_2024}. Given that their habitats and migratory patterns cover large expanses of remote areas, tracking commonly uses crewed aircraft \cite{Moll:2022aa}. These require highly-trained pilots accepting of risk, airports and maintenance infrastructure. Additionally, these crafts increase pollution, both greenhouse gas and noise. The use of small autonomous drones can address these challenges \cite{gentle_comparison_2018}. A drone system that can independently and efficiently navigate, locate and identify the wildlife likely requires a system that can, using the standard photographic and limited computational equipment onboard, detect the objects below \cite{Zhou:2022aa}. Ke et al. \cite{ke_deep_2024} recently demonstrated that a lightweight filter-detect-classify pipeline can run entirely in-flight, edging toward real-time wildlife surveys on embedded GPUs. Their binary-classifier "image filter" eliminated >80\% of empty frames before detection, cutting on-board compute without harming abundance estimates.

Tracking moose, as this paper describes, is a problem within aerial small-object detection using UAVs \cite{Hua:2025aa}.
The challenge for the model is to find small, low-resolution objects \cite{kassim2020small, gowravaram2023prescribed} viewed at an angle which is unusual to standard object detection datasets \cite{lin_microsoft_2015, everingham_pascal_2010}. In the wildlife case, the surrounding environment may be cluttered and the fauna may be occluded by objects or the edge of the frame \cite{risse2017visual}. Natural footage results in varying conditions such as lighting, motion blur and altitude \cite{iglay2024wildlife, HodgsonJarrodC.2018Dcwm, CorcoranEvangeline2021Adow}. Therefore, an effective system will have been exposed to the various cases during the training process.

This paper includes a comparative study of efficient and advanced models from three object detection architecture families---YOLOv11 \cite{Jocher_Ultralytics_YOLO_2023}, Faster R-CNN \cite{ren_faster_2017}, and Co-DETR \cite{zong_detrs_2023}---for aerial wildlife object detection. 
This study applies a multi-scale patching method we employed for other applications in prior works \cite{rahman2024new,Rahman:2023aa}, as visualized in Fig.~\ref{fig:patchdiagram}. This method limits the pixel quantity sent to the model to decrease computational complexity, exposes the model to the objects at multiple scales and studies performance on partial instances which were occluded by the image edge.
Current gaps and future research directions for improving wildlife detection from drones are discussed afterwards.

\section{Related Works}
\label{sec:relatedworks}

YOLO-based \cite{redmon_you_2016} models are fast in general due to their one-stage (localization and classification in a single pass) object detection algorithm. They have been used extensively in wildlife aerial object detection, commended for their balance of efficiency and accuracy \cite{ramesh_improvements_2023, leng_recent_2024, wang_research_2021, atik_comparison_2022, hu_efficient-lightweight_2023, zhang2020efficient, ma_yolo-uav_2023}. Studies citing YOLOv11 \cite{Jocher_Ultralytics_YOLO_2023} for aerial object detection include: ship detection \cite{ji_mbsdet_2024,li_oriented_2025}, though ships are inflexible unlike a moose; photovoltaic flaws \cite{naeem_aerial_2025,wang_cpdd_2025}, though these lack the variation of natural scenes; others focus on IR and other modalities \cite{yang_beyond_2025,wang_real-time_2025}, whereas ours assumes only RGB.

Though Faster R-CNN \cite{ren_faster_2017} has been succeeded by the likes of Mask R-CNN \cite{He_2017_ICCV}, Mask R-CNN requires additional computations for pixel-level segmentation. In a comparison with the later Libra R-CNN \cite{Pang:2019aa} and RetinaNet \cite{lin_focal_2018}, all three algorithms showed poor performance when objects were tightly grouped. However, Faster R-CNN remained competitive \cite{delplanque_multispecies_2022} and is worth including as it has and continues to be used extensively as a base model \cite{wang_optimized_2023, huang_improved_2022} or in studies \cite{ezzeddini_analysis_2024}.

Transformer-based object detection architectures are strongly represented among the very best object detectors, since their self-attention mechanism permits the global context which is useful for considering the entire image and contextual relationships within it \cite{chen2023accumulated, zhu_deformable_2021, zhang2025depth}. This comes at the expense of efficiency, so employing a transformer model on a drone is challenging \cite{lu_lightweight_2024}. As of July 2025, detection transformer Co-DETR \cite{zong_detrs_2023} remains one of the very best models for detection, as listed on PapersWithCode \cite{detr_webpage}.

{\bf Impact of image patch size and augmentation techniques:}
Objects in aerial object detection may be small; as methods frequently resize the original image prior to processing, those small objects might be reduced to only a few pixels \cite{Shin:2024aa}. One potential solution is to carve the image into "patches", regions within the image each treated as a new image. As this multiplies the number of pixels requiring processing by the number of patches, it is paramount that the associated model be fast. Choosing an appropriate image patch size for processing can significantly affect detection performance \cite{liu_esod_2024}. Large images can be split into smaller patches (tiles or "chips") to zoom in on details; however, too-small patches may lose contextual cues and cut objects at boundaries. Researchers commonly experiment with different patch sizes or tiling strategies in aerial detection. For example, the VisDrone \cite{zhu2021detection} dataset uses image chips for training detectors, and Hong et al. \cite{hong_patch-level_2019} generated hard example chips by cutting and pasting objects into new backgrounds to improve small object detection. Generally, smaller patches (higher relative scale) help in detecting small targets because the object appears larger to the detector; analogous to using a higher input resolution. Though small patches can reveal small objects that would be missed when downsampled, very small patches or minimal tiling means drastically more computation and more false positives from noise-generated background blobs.

A common strategy is to use multi-scale training and augmentation so that the model sees objects at various scales. Copy-paste augmentation places objects from one image into another. The patch-level augmentation method by Hong et al. uses the idea that copy-and-pasting of annotations can combat class imbalance \cite{hong_patch-level_2019}. However, if data is limited as in the moose dataset, bootstrapping in this way may not provide much additional variation. 

Therefore, this paper's contributions are situated as follows:
\begin{itemize}
    \item To the best of the authors' knowledge at time of publication, this is the only aerial object detection study (and certainly the only one for moose in arctic settings specifically) which directly compares: 
    \begin{itemize}
        \item YOLOv11 \cite{khanam_yolov11_2024} (YOLO family, one-stage convolutional, recent)
        \item Faster R-CNN \cite{ren_faster_2017} (R-CNN family, two-stage convolutional, more classic)
        \item Co-DETR \cite{zong_detrs_2023} (DETR family, transformer-based)
    \end{itemize}
    \item This paper provides an inspection of the qualities of a dataset (sourced from four datasets \cite{silent_wings_moose_nodate, silent_wings_moose_nodate-1, silent_wings_moose_nodate-2, silent_wings_moose_nodate-3}) of arctic moose images, resulting from our patching scheme, using in-depth statistical analyses of model behaviour at various settings.
    \item We describe the patching scheme as a model-agnostic means of training models on different scales while reducing the number of pixels sent through the model at one time; even if only tested on large patches, results remain strong.
\end{itemize}


\section{Dataset}
\label{sec:dataset}
The moose aerial object detection dataset, used for finetuning and evaluating, is assembled from four publicly available datasets \cite{silent_wings_moose_nodate, silent_wings_moose_nodate-1, silent_wings_moose_nodate-2, silent_wings_moose_nodate-3}. In total, the 1694 full-size images in the dataset are all aerial photos taken during the day of moose in a snowy environment. "Moose" is the single annotated class, for which non-oriented bounding boxes are provided as annotations. Four sizes of original images exist; $(4000,6000), (6000,4000)$ and $(1365,2048), (2048, 1365)$. Since the moose appear in arbitrary orientations, the image dimensions can be switched freely; therefore, Table~\ref{table:imagesizes} consolidates portrait and landscape. If the original images were processed verbatim, resizing could reduce smaller annotations to just a few pixels, making them extremely difficult to detect.

To address this, multiple patch sizes were incorporated during training to help the model learn to recognize objects at different scales. Table~\ref{table:imagesizes} also specifies, given the size of the original image, the size of the patches created from it. The scale factors were carefully selected so that the dimensions of large, medium, and small patches remain consistent across varying image sizes. This normalization ensures that both patch and annotation sizes are treated fairly when resized by the model, promoting balanced learning across different object scales.

Table~\ref{table:datastatistics} showcases some statistics of the (threshold = 0.5, overlap = 0.1 "default") dataset, once each original image has been dissected into patches. The bounding boxes are tiny and sparse; many patches contain no annotations, and these were unused, leaving 1771 patches. As such, the training speed does not decrease significantly; in fact, many fewer pixels are used. That said, all collected imagery can (and should) be retained so that post-flight audits can quantify false negatives and verify conservation-grade accuracy \cite{ke_deep_2024}. As the large patches have only slightly more annotations (moose) on average than the small patches, this indicates the sparseness of the moose in the images. The \textbf{avg. BBox area (norm.)} in Table~\ref{table:datastatistics} indicates average bounding box size relative to average patch size, computed as the \textbf{avg. BBox area (norm.)} divided by the average size of patches of that size (and rounded). Through patch scaling, the result is that the average size of bounding boxes relative to the patch becomes significantly more balanced.

\begingroup                 
\setlength{\tabcolsep}{4pt} 
\begin{table}[t]
\begin{center}
\begin{tabular}{|c|c|c|c|c|c|c|c|c|c|}
\hline
\multicolumn{2}{|c|}{~} & \multicolumn{8}{c|}{\textbf{Original Image Sizes}} \\
\cline{3-10}
\multicolumn{2}{|c|}{~} & \multicolumn{4}{c|}{(4000, 6000)*} & \multicolumn{4}{c|}{(1365, 2048)*} \\
\cline{2-10}
~ & \multicolumn{1}{c|}{\textbf{Overlap}} & 0.0 & 0.1 & 0.3 & \textbf{Scale} & 0.0 & 0.1 & 0.3 & \textbf{Scale} \\
\hline
\cline{3-10}
\multirow{3}{*}{\rotatebox[origin=c]{90}{\textbf{Patch Size~}}} & \addstackgap[5pt]{Large} & \multicolumn{1}{|c|}{(667, 1000)} & (722, 1083) & (833, 1250) & 1/6 & (682, 1024) & (716, 1075) & (784, 1178) & 1/2 \\
\cline{2-10}
& \addstackgap[5pt]{Medium} & \multicolumn{1}{|c|}{(500, 750)} & (544, 816) & (631, 947) & 1/8 & (455, 683) & (485, 728) & (546, 819) & 1/3 \\
\cline{2-10}
& \addstackgap[5pt]{Small} & \multicolumn{1}{|c|}{(333, 500)} & (364, 546) & (425, 638) & 1/12 & (341, 512) & (367, 550) & (418, 627) & 1/4 \\
\hline
\end{tabular}
\end{center}
\caption{The dimensions of the original images and the generated patches. Scales were chosen such that each ``Large", ``Medium" and ``Small" patch would be similar in size. \label{table:imagesizes}}
\begin{center}
\vspace{-10pt}
\footnotesize{* Includes images of size $(x,y)$ and $(y,x)$; those in the latter are rotated.}
\end{center}
\end{table}
\endgroup

\begin{table}[t]
\begin{center}
\begin{tabular}{|c|c|c|c|c|c|}
\hline
\textbf{Split} & \textbf{Patch Sizes} & \textbf{\# Images} & \textbf{avg. annos/img} & \textbf{avg. BBox area} & \textbf{avg. BBox area (norm.)} \\
\hline
train & all & 1408 & 1.72 & 7019 & 0.013 \\
val & all & 363 & 1.74 & 10178 & 0.017 \\
\hline
train & large & 440 & 1.87 & 11871 & 0.02 \\
train & medium & 469 & 1.72 & 7298 & 0.02 \\
train & small & 499 & 1.58 & 1671 & 0.01 \\
\hline
val & large & 124 & 1.80 & 20957 & 0.03 \\
val & medium & 116 & 1.76 & 7157 & 0.02 \\
val & small & 123 & 1.67 & 1501 & 0.01 \\
\hline
\end{tabular}
\end{center}
\caption{The statistics of the "default" (threshold = 0.5, overlap = 0.1) version of the dataset, representing the "middle" for both threshold and overlap. \label{table:datastatistics}}
\end{table}

As the image is divided into patches, it is possible that an annotation may fall on the border of the patch, and so the model must learn and understand the small object based on a small portion of it \cite{rahman2024new}. We set and experiment on a threshold governing the proportion of an annotation that is cut off, above which that annotation is not used for that patch. Though it could be argued that learning to identify partial annotations is useful for guiding the drone, the likelihood that the object appears at the edge of the original image is somewhat low. The object may appear at the edge of a patch, and for this we introduce the second hyperparameter, the overlap. This parameter controls how much the patches overlap, and depending on its value, it can help ensure that the annotation appears in its complete form in at least one patch for the purpose of training and inference. Overlap does adjust the patch size, as shown in Table~\ref{table:imagesizes}. Model performance on dataset versions governed by these hyperparameters is explored in Appendices~\ref{appendix:experimentstable},\ref{appendix:statisticalanalysis}. The joint effect of patch size and overlap on the patch dimensions (and therefore on the performance metrics) is discussed in Appendices~\ref{appendix:YOLOv11DataTable},\ref{appendix:YOLOpatchsize}.


\section{Models}
\label{sec:models}

YOLOv11 \cite{Jocher_Ultralytics_YOLO_2023} is the newest iteration of the "You Only Look Once" one-stage detectors. It builds on the YOLO series’ emphasis on real-time performance, introducing architectural innovations for higher accuracy without sacrificing speed. YOLOv11 introduces the C2PSA (Cross Stage Partial with Spatial Attention) module which leverages attention to improve focus on key areas (including objects). Its C3k2 block is a more efficient version of the CSP (Cross Stage Partial) \cite{bochkovskiy_yolov4_2020}, the latter introduced to ensure that the same gradients are computed multiple times through a deep network. To further speed up processing, SPPF (Spatial Pyramid Pooling - Fast) is used as pooling. Combined, these permit YOLOv11 to exceed the mAP of its YOLO predecessors with fewer parameters \cite{jegham_evaluating_2024}. YOLO models use multi-scale feature maps and dense predictions, which can identify small objects well assuming good training.

Faster R-CNN (Region-based Convolutional Neural Network) is a classic two-stage detector that has been widely used since its introduction by Ren et al. \cite{ren_faster_2017, Delplanque:2024aa}. Its main contribution is the Region Proposal Network (RPN), in which the main network generating feature maps for the region-based detectors only needs to be supplemented with two small layers so that a number of objectness scores and bounds are proposed for each location. Default proposal anchors are as small as $128 \times 128$ pixels, and after stride-16 down-sampling this corresponds to objects only $8 \times 8$ pixels on the original image; the domain of small object detection. Two stage models generate region proposals (performing the easier task of "object or no object") and then classify and refine them, which often yields higher precision at the cost of speed \cite{redmon_you_2016}. Nevertheless, Faster R-CNN when released was state-of-the-art in both detection accuracy and speed (5 fps, nearly real-time) \cite{ren_faster_2017}. 

Co-DETR (Collaborative DETR) \cite{zong_detrs_2023} is a recent transformer-based object detector that builds on the DETR \cite{carion_end--end_2020} paradigm with innovative training schemes for improved efficiency. DETR models do end-to-end object detection by dispensing with proposal generation and non-maximum suppression, instead using a set-based global loss and transformers to directly predict objects. 
Previous DETRs used one-to-one set matching (one query per ground truth); with fewer positive samples, training becomes unstable and slow \cite{zong_detrs_2023}. The namesake collaborative auxiliary heads are discarded before inference, so there is no additional computational cost. A multi-scale adapter accommodates features at various scales.
The result is a detector that achieves cutting-edge accuracy on standard benchmarks, using fewer parameters than similarly-performing competitors. 
Although Co-DETR is computationally heavy (inference is slower than YOLO; large models are not real-time), it provides a high-accuracy benchmark. 
This comparative perspective is valuable, as initial evidence suggests each approach has different strengths: YOLOv11 prioritizes speed, Faster R-CNN precision, and Co-DETR raw accuracy. 

\section{Experiments}
\label{sec:experments}

\subsection{Model Performance by Threshold-Overlap Values}

Naturally, the first significant hyperparameter in each of our experiments is the selected model. From the aforementioned YOLOv11, Faster R-CNN and Co-DETR, the default settings of Ultralytics \cite{Jocher_Ultralytics_YOLO_2023} and MMDetection \cite{MMDetection_Contributors_OpenMMLab_Detection_Toolbox_2018} are used. In the interest of efficiency, we use the Co-DETR that includes ResNet50 \cite{he_deep_2016} rather than SWIN \cite{zong_detrs_2023}. All experiments were conducted on 1 NVIDIA A100 80GB, which is significantly more powerful than what may be expected of a drone, but architecture was chosen in each case to minimize the computational complexity.

The train and test datasets are used completely, with no biasing of patch size represented in each epoch. Convergence is fast, as (by the limitations of the dataset) the lighting is constant and there is only one class. For Faster R-CNN and Co-DETR, a mere 5 epochs were able to provide a good result, with the best epoch kept.

For the dataset, the following two hyperparameters are varied in typical Grid Search fashion:
\begin{itemize}
    \item \verb|Threshold.| Values: $T \in \{0.1, 0.3, 0.5, 1.0\}$. Dividing the image into patches results in more borders; the patch edge may lead to partial moose instances. For a set threshold $T \in [0, 1]$, annotations where the proportion of the bounding box that is on the image (not cut off by the edge) is less than $T$ are discarded and treated as ground-truth background. This hyperparameter permits study of model performance on partial examples, where the small-object detection task becomes even more challenging. Effective models must learn useful features at tiny scales to be effective here. In the context of aerial detection, partial instances are important to consider since a moose might appear at the border of the UAV's field of view. At $T = 1.0$, an annotation is kept only if it is not occluded at all by the patch edge (but the moose itself may be occluded by image contents).
    \item \verb|Overlap.| Values: $O \in \{0.0, 0.1, 0.3\}$. An image can be divided into patches by simply cutting. As discussed in \verb|Threshold| above, this cut may bifurcate image instances. If simple cutting is applied, a particular instance may only appear as a partial instance at a given scale. With an overlap, objects that appear partially in one patch will fully appear in an adjacent patch, ensuring that both a whole and partial instance of those moose exist in the training set. By doing this, features become more resilient. For overlap value $O \in [0, 1]$, this is accomplished by making the patches slightly larger, but not so much larger so as to interfere with the patch sizes. Simple cutting is the case $O = 0.0$. 
\end{itemize}

\begin{figure}[t]
  \centering
  \begin{subfigure}[b]{0.48\textwidth}
    \centering
    \includegraphics[scale=0.47]{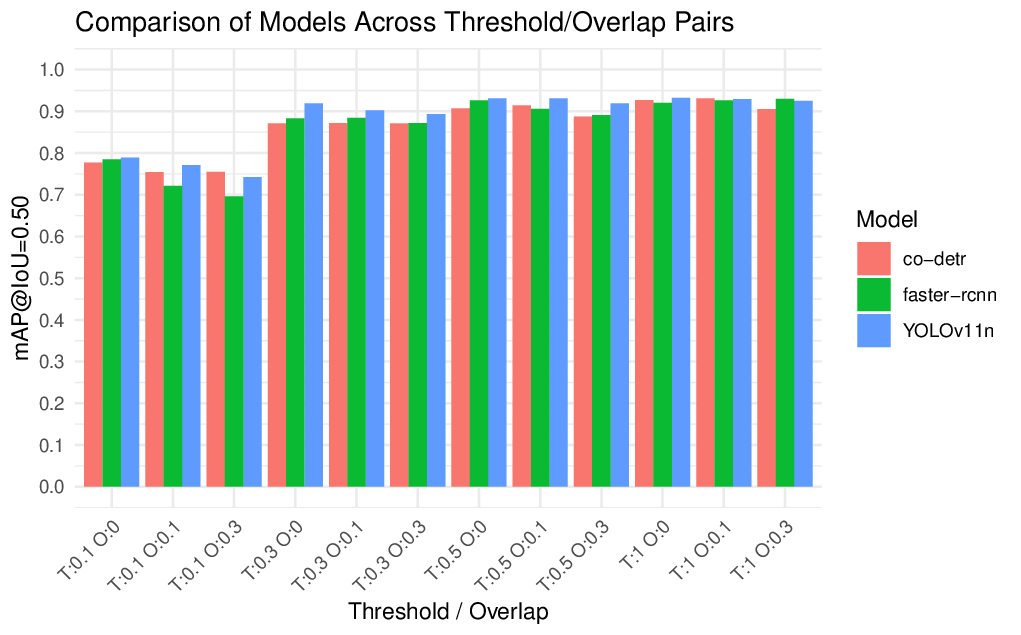}
    \caption{mAP@IoU=0.50}
    \label{fig:barchart50}
  \end{subfigure}\hfill
  \begin{subfigure}[b]{0.48\textwidth}
    \centering
    \includegraphics[scale=0.47]{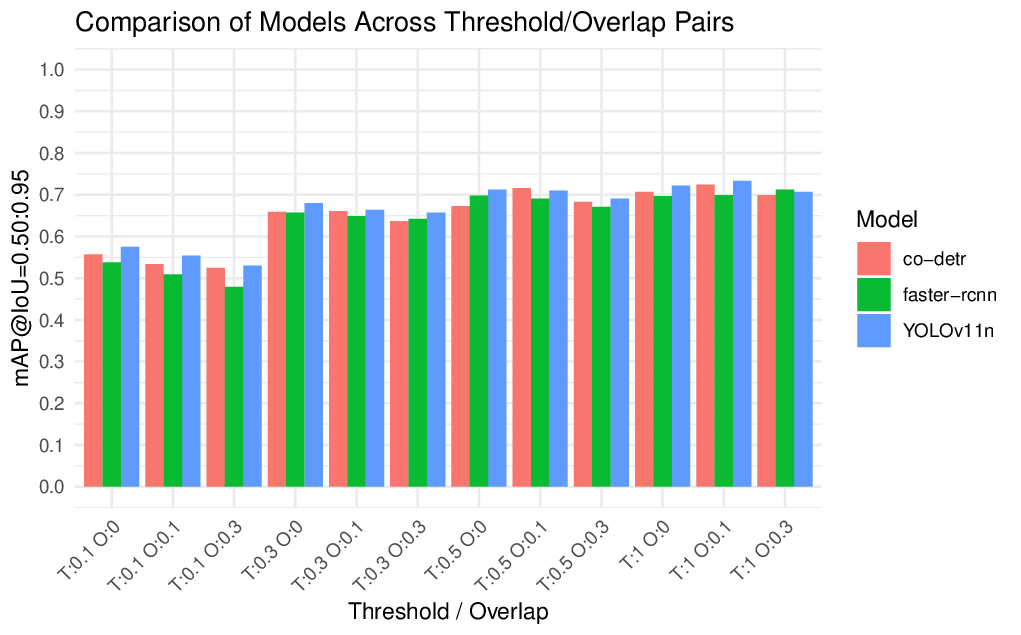}
    \caption{mAP@IoU=0.50:0.95}
    \label{fig:barchart95}
  \end{subfigure}

  \vspace{1em} 

  \begin{subfigure}[b]{0.48\textwidth}
    \centering
    \tiny
    \begin{verbatim}
Call:
lm(formula = IoU.0.50 ~ model + poly(Threshold, 3) * Overlap, 
    data = data)

Residuals:
      Min        1Q    Median        3Q       Max 
-0.031270 -0.007877  0.000081  0.005239  0.032766 

Coefficients:
                             Estimate Std. Error t value Pr(>|t|)    
(Intercept)                  0.874401   0.004572 191.269  < 2e-16 ***
modelfaster-rcnn            -0.002583   0.005502  -0.470 0.642585    
modelYOLOv11n                0.017667   0.005502   3.211 0.003504 ** 
poly(Threshold, 3)1          0.284602   0.019727  14.427 6.39e-14 ***
poly(Threshold, 3)2         -0.225377   0.019727 -11.425 1.24e-11 ***
poly(Threshold, 3)3          0.059362   0.019727   3.009 0.005756 ** 
Overlap                     -0.076131   0.018008  -4.228 0.000258 ***
poly(Threshold, 3)1:Overlap  0.240274   0.108050   2.224 0.035064 *  
poly(Threshold, 3)2:Overlap -0.125353   0.108050  -1.160 0.256533    
poly(Threshold, 3)3:Overlap  0.173941   0.108050   1.610 0.119512    
---
Signif. codes:  0 ‘***’ 0.001 ‘**’ 0.01 ‘*’ 0.05 ‘.’ 0.1 ‘ ’ 1

Residual standard error: 0.01348 on 26 degrees of freedom
Multiple R-squared:  0.9735,	Adjusted R-squared:  0.9643 
F-statistic: 106.1 on 9 and 26 DF,  p-value: < 2.2e-16
    \end{verbatim}
    \caption{Linear fit summary, mAP@IoU=0.50}
    \label{fig:linreg50}
  \end{subfigure}\hfill
  \begin{subfigure}[b]{0.48\textwidth}
    \centering
    \tiny
    \begin{verbatim}
Call:
lm(formula = IoU.0.50.0.95 ~ model + poly(Threshold, 3) * Overlap, 
    data = data)

Residuals:
       Min         1Q     Median         3Q        Max 
-0.0272500 -0.0053690 -0.0008036  0.0073899  0.0210833 

Coefficients:
                             Estimate Std. Error t value Pr(>|t|)    
(Intercept)                  0.657274   0.003975 165.355  < 2e-16 ***
modelfaster-rcnn            -0.011083   0.004784  -2.317 0.028647 *  
modelYOLOv11n                0.013333   0.004784   2.787 0.009799 ** 
poly(Threshold, 3)1          0.306630   0.017152  17.877 3.98e-16 ***
poly(Threshold, 3)2         -0.218056   0.017152 -12.713 1.15e-12 ***
poly(Threshold, 3)3          0.053288   0.017152   3.107 0.004535 ** 
Overlap                     -0.070179   0.015658  -4.482 0.000132 ***
poly(Threshold, 3)1:Overlap  0.249777   0.093948   2.659 0.013246 *  
poly(Threshold, 3)2:Overlap -0.110337   0.093948  -1.174 0.250862    
poly(Threshold, 3)3:Overlap  0.060424   0.093948   0.643 0.525752    
---
Signif. codes:  0 ‘***’ 0.001 ‘**’ 0.01 ‘*’ 0.05 ‘.’ 0.1 ‘ ’ 1

Residual standard error: 0.01172 on 26 degrees of freedom
Multiple R-squared:  0.9807,	Adjusted R-squared:  0.974 
F-statistic: 146.5 on 9 and 26 DF,  p-value: < 2.2e-16
    \end{verbatim}
    \caption{Linear fit summary, mAP@IoU=0.50:0.95}
    \label{fig:linreg95}
  \end{subfigure}

  \caption{Summary of experimental results comparing mAP@IoU=0.5 and mAP@IoU=0.5:0.95 for Model-Threshold-Overlap triples. Fig.~\ref{fig:barchart50} presents the data for IoU=0.5 which Fig.~\ref{fig:linreg50} fits with a linear model; Fig.~\ref{fig:barchart95} and Fig.~\ref{fig:linreg95} do the same for IoU=0.5:0.95. In short, the model, threshold and overlap all significantly affect mAP. See Appendix~\ref{appendix:experimentstable} for the data and Appendix~\ref{appendix:statisticalanalysis} for statistical justification.}
  \label{fig:combined}
\end{figure}

This study carried out a number of experiments, varying the model, threshold and overlap as explained above. Two standard metrics are used for evaluation on the test portion of the dataset; mAP@IoU=0.5 and mAP@IoU=0.5:0.95. IoU=$X$ enforces that each predicted bounding box's overlap with the ground truth box must be at least $100 \times X$\% of the area of the union of those boxes for the prediction to be marked as correct. The metric mAP@IoU=0.5:0.95 is the average of mAP@IoU=$X$, for all $X \in \{0.5, 0.55, \ldots 0.9, 0.95\}$. In all cases, mAP@IoU=0.5 $\geq$ mAP@IoU=0.5:0.95 since the latter has a higher average overlap requirement. These have been recorded in Appendix~\ref{appendix:experimentstable} and analyzed in Fig.~\ref{fig:combined} (visualized in Figs.~\ref{fig:barchart50},~\ref{fig:barchart95}, modeled in Figs.~\ref{fig:linreg50},~\ref{fig:linreg95}). The bounding-box annotations and associated score operations are adjusted as a result of the patch cropping, since the score metrics are functions of IoU.

To analyze the effects of the grid search hyperparameters, this study includes a linear regression analysis in Figs.~\ref{fig:linreg50},~\ref{fig:linreg95}. First, analysis of the models; represented as a categorical variable with possible values \{\verb|modelCo-DETR, modelfaster-rcnn, modelYOLOv11n|\}. \verb|modelCo-DETR| does not appear in the analysis; alphabetically first, it is assigned as a default value and is normalized to have an effect of 0. The coefficient on \verb|modelfaster-rcnn| is negative and small, whereas the coefficient on \verb|modelYOLOv11n| is positive and small. This suggests that throughout the experiments, generally YOLOv11 $>$ Co-DETR $>$ Faster R-CNN in terms of performance, but they are all similar; this is verified in Table~\ref{table:results}. Worse models seemed to prefer higher overlap values, perhaps since seeing full moose examples was necessary and outweighed the extra non-moose pixels. Faster R-CNN, being both older than YOLOv11 and less complex than Co-DETR, is expected to finish in third place. The fact that the model performance does not differ much is reiterated by the $p$-value, \verb|Pr(>t)|, which is greater than $0.05$ for \verb|modelfaster-rcnn| and \verb|modelYOLOv11n|; changing the model does not have a statistically-significant effect. One is therefore inclined to believe that for this dataset, the model choice among these three is not important for mAP. For the purpose of UAV deployment, speed and memory requirements are shown in Table~\ref{table:model_stats} to contextualize the results. Despite all having similar peak mAP@IoU=0.5, YOLOv11 was by far the smallest, lightest, fastest and most memory-friendly.

The threshold hyperparameter, represented as a cubic function \verb|poly(Threshold, 3)|, has all of its coefficients significant. That its leading coefficient is positive suggests that threshold value should be positively correlated with mAP. This may seem like a trivial observation; discarding partial moose instances would make the task easier. However, based on Figs.~\ref{fig:barchart50},~\ref{fig:barchart95} all of $T \in \{0.3, 0.5, 1.0\}$ seem to be similar, only $T = 0.1$ is noticeably worse. This suggests that the models can be trained to recognize only 30\% of a moose without much difficulty. Additionally, the fact that the threshold was better represented by a cubic than as a standard linear variable (Appendix~\ref{appendix:statisticalanalysis}) suggests a more complex relationship. This can be explained by the diminishing returns for higher thresholds; raising the threshold from 0.1 to 0.2 would (theoretically) lead to large mAP gains since bounding boxes which are less than 20\% visible are extremely difficult for even a top model to discover, but from 0.9 to 1.0 would likely make little difference since the models are strong enough to deal with mild occlusions.

Overlap also appears to have a significant effect on performance, with the correlation with accuracy negative based on the bar graphs and coefficient estimates. The negative effect seems to be stronger at low threshold values, as corroborated by the significant interaction term \verb|poly(Threshold, 3)1:Overlap|. This may be a consequence of the overlap increasing the patch sizes, thereby reducing the magnification on the tiny object fragments. Increasing overlap improves mAP by showing whole bounding boxes but decreases mAP since the larger image results in fewer pixels for each instance, a relationship which is captured by the \verb|Overlap| terms.

\begin{table}[t]
\begin{center}
\begin{tabular}{|c|c|c|c|c|c|c|c|c|}
\hline
\textbf{~} & \multicolumn{4}{c|}{mAP@IoU=0.5 (Fig.~\ref{fig:barchart50})} & \multicolumn{4}{c|}{mAP@IoU=0.5:0.95 (Fig.~\ref{fig:barchart95})} \\
\hline
\textbf{Model} & \textbf{Average} & \textbf{Max} & \textbf{T} & \textbf{O} & \textbf{Average} & \textbf{Max} & \textbf{T} & \textbf{O} \\
\hline
Faster R-CNN\cite{ren_faster_2017} & 86.4\% & 93.0\% & 1.0 & 0.3 & 63.7\% & 71.2\% & 1.0 & 0.3 \\
\hline
Co-DETR\cite{zong_detrs_2023} & 86.4\% & 93.1\% & 1.0 & 0.1 & 64.5\% & 72.4\% & 1.0 & 0.1 \\
\hline
YOLOv11\cite{Jocher_Ultralytics_YOLO_2023} & \textbf{88.2\%} & \textbf{93.2\%} & 1.0 & 0.0 & \textbf{66.1\%} & \textbf{73.3\%} & 1.0 & 0.1 \\
\hline
\end{tabular}
\end{center}
\caption{Summary of experimental results. For each model, \textbf{Average} is the average performance on all experiments (all threshold/overlap pairs), \textbf{Max} is the maximum of those, \textbf{T} and \textbf{O} are the threshold and overlap settings for that maximum. \label{table:results}}
\begin{center}
\vspace{-10pt}
\end{center}
\end{table}

\begingroup                 
\setlength{\tabcolsep}{3pt} 
\renewcommand{\arraystretch}{1.1} 

\begin{table}[t]
\centering
\begin{tabular}{@{}|c|c|c|c|c|c|c|c|c|c|@{}} 
\hline
\textbf{Model} & \textbf{Layers} & \textbf{Params} & \textbf{Size} &
\textbf{sec/iter} & \textbf{GFLOPs} & \textbf{Img Dim.} &
\textbf{Batch} & \textbf{Peak RAM} & \textbf{mAP@0.5} \\
\hline
Faster R-CNN\cite{ren_faster_2017} & 223 & 64M & 758\,MB & 0.35 & 418 & 800$\times$1216 & 4 & 16 & 93.0 \\
\hline
Co-DETR\cite{zong_detrs_2023} & 279 & 99M & 790\,MB & 0.81 & 860 & 800$\times$1200 & 1 & 16 & 93.1 \\
\hline
YOLOv11n\cite{Jocher_Ultralytics_YOLO_2023} & 181 & 2.6M & \textbf{5.22\,MB} & \textbf{0.17} & \textbf{6.4} & 640$\times$640 & 16 & \textbf{3.04} & \textbf{93.2} \\
\hline
\end{tabular}
\vspace{8pt}
\caption{The models and their performances, contextualized with speed and hardware requirements.}
\label{table:model_stats}
\end{table}

\endgroup                   


\subsection{Assessing Data Usage and Per-Size Performance}

Using the winning YOLOv11 model, some other experiments were run to strengthen the evaluation of the patching method (refer to Appendix~\ref{appendix:YOLOv11DataTable} for experiments and Appendix~\ref{appendix:YOLOpatchsize} for methods). Firstly, for each threshold-overlap pair for the purpose of this section, YOLOv11 was trained on the entire training set of that threshold-overlap version (as were the other models for all experiments) and was then evaluated separately on the small, medium and large subsets of the associated validation set. This permits an exploration of performance on differently-sized patches (moose of different clarity). Size was a significant factor, with the small patches yielding the best performance. As YOLOv11 resizes patches, no non-noise difference in per-patch inference speed was noted. The number of training images was found to have a small but significant negative correlation with mAP, perhaps owing to the large negative regions.

\section{Conclusion}

To the best of the authors' knowledge at time of publication, this is the first study to compare YOLOv11, Faster R-CNN and Co-DETR directly for aerial object detection. The three models are applied to wildlife object detection on a dataset of moose in arctic scenes. By comparing models, thresholds and overlap factors, this paper has shown that good results can be attained on models and patch settings which are reasonable for UAV usage, even if only some patch scales get used for memory or speed reasons. 

Three models were compared, of different ages and architectures: YOLOv11 is a fast and recent one-stage convolutional system; Faster R-CNN is an older yet still lightweight two-stage convolutional system that may benefit most from patch overlapping; Co-DETR is recent and a transformer, although one that has been modified to improve performance without additional parameters at inference time. All models performed similarly on nearly all configurations of the dataset (Figs.~\ref{fig:barchart50},~\ref{fig:barchart95}), though perhaps YOLOv11 $>$ Co-DETR $>$ Faster R-CNN. For UAV usage, efficiency is also considered; YOLOv11 would be the best choice for performance and speed, and then users could choose Co-DETR or Faster R-CNN based on UAV hardware availability.

For future work, there are numerous opportunities for architectural enhancements. For models like Co-DETR and other transformer-based approaches, incorporating a focal attention mechanism may improve the detection of small, low-contrast objects and reduce reliance on small input patches. As we expand the dataset to include more diverse conditions—such as varying lighting—models can be further evaluated and fine-tuned for robustness. To serve as an improver of performance, a more sophisticated patching method that balances the positive and negative regions could be applied.

\section*{Acknowledgements}
This research was funded by the Natural Sciences and Engineering Research Council of Canada (NSERC) under grant numbers ALLRP 588173-23 and ALLRP 570580-21.

\section*{Abbreviations}
\begin{tabular}{@{}p{0.1\linewidth}p{0.9\linewidth}}
BBox & Bounding Box \\
UAV & Uncrewed Aerial Vehicle; Drone
\end{tabular}

\bibliographystyle{myIEEEtran}
\bibliography{references}  

\newpage

\appendix
\section*{Appendix}

\section{Experiments Table}
\label{appendix:experimentstable}

\begin{table}[!htbp]
\begin{center}
\begin{tabular}{|c|c|c|c|c|c|}
\hline
\textbf{Model*} & \textbf{Threshold} & \textbf{Overlap} & \textbf{sec/iter} & \textbf{IoU=0.50:0.95} & \textbf{IoU=0.50} \\
\hline
YOLOv11n & 0.1 & 0.0 & 0.2519 & 0.575 & 0.789 \\
\hline
YOLOv11n & 0.1 & 0.1 & 0.3333 & 0.554 & 0.771 \\
\hline
YOLOv11n & 0.1 & 0.3 & 0.6944 & 0.530 & 0.742 \\
\hline
YOLOv11n & 0.3 & 0.0 & 0.1553 & 0.680 & 0.919 \\
\hline
YOLOv11n & 0.3 & 0.1 & 0.1634 & 0.664 & 0.902 \\
\hline
YOLOv11n & 0.3 & 0.3 & 0.2841 & 0.657 & 0.893 \\
\hline
YOLOv11n & 0.5 & 0.0 & 0.1776 & 0.712 & 0.931 \\
\hline
YOLOv11n & 0.5 & 0.1 & 0.1645 & 0.710 & 0.931 \\
\hline
YOLOv11n & 0.5 & 0.3 & 0.1555 & 0.691 & 0.919 \\
\hline
YOLOv11n & 1.0 & 0.0 & 0.1621 & 0.722 & 0.932 \\
\hline
YOLOv11n & 1.0 & 0.1 & 0.1721 & 0.720 & 0.929 \\
\hline
YOLOv11n & 1.0 & 0.3 & 0.1597 & 0.707 & 0.925 \\
\hline
Co-DETR & 0.1 & 0.0 & 0.8118 & 0.557 & 0.777 \\
\hline
Co-DETR & 0.1 & 0.1 & 0.8208 & 0.534 & 0.754 \\
\hline
Co-DETR & 0.1 & 0.3 & 0.8423 & 0.525 & 0.755 \\
\hline
Co-DETR & 0.3 & 0.0 & 0.8223 & 0.659 & 0.871 \\
\hline
Co-DETR & 0.3 & 0.1 & 0.8138 & 0.661 & 0.872 \\
\hline
Co-DETR & 0.3 & 0.3 & 0.8334 & 0.637 & 0.871 \\
\hline
Co-DETR & 0.5 & 0.0 & 0.8200 & 0.673 & 0.907 \\
\hline
Co-DETR & 0.5 & 0.1 & 0.8113 & 0.716 & 0.914 \\
\hline
Co-DETR & 0.5 & 0.3 & 0.8064 & 0.683 & 0.887 \\
\hline
Co-DETR & 1.0 & 0.0 & 0.8173 & 0.707 & 0.927 \\
\hline
Co-DETR & 1.0 & 0.1 & 0.8177 & 0.724 & 0.931 \\
\hline
Co-DETR & 1.0 & 0.3 & 0.8281 & 0.699 & 0.905 \\
\hline
Faster R-CNN & 0.1 & 0.0 & 0.3580 & 0.538 & 0.785 \\
\hline
Faster R-CNN & 0.1 & 0.1 & 0.3584 & 0.509 & 0.721 \\
\hline
Faster R-CNN & 0.1 & 0.3 & 0.3605 & 0.479 & 0.696 \\
\hline
Faster R-CNN & 0.3 & 0.0 & 0.3592 & 0.657 & 0.883 \\
\hline
Faster R-CNN & 0.3 & 0.1 & 0.3597 & 0.649 & 0.884 \\
\hline
Faster R-CNN & 0.3 & 0.3 & 0.3592 & 0.642 & 0.872 \\
\hline
Faster R-CNN & 0.5 & 0.0 & 0.3585 & 0.698 & 0.926 \\
\hline
Faster R-CNN & 0.5 & 0.1 & 0.3569 & 0.691 & 0.906 \\
\hline
Faster R-CNN & 0.5 & 0.3 & 0.3590 & 0.671 & 0.891 \\
\hline
Faster R-CNN & 1.0 & 0.0 & 0.3607 & 0.697 & 0.920 \\
\hline
Faster R-CNN & 1.0 & 0.1 & 0.3594 & 0.699 & 0.926 \\
\hline
Faster R-CNN & 1.0 & 0.3 & 0.3581 & 0.712 & 0.930 \\
\hline
\end{tabular}
\vspace{10pt}
\caption{Experiment data. Speed (sec/iter) and accuracy (mAP) metrics across models, thresholds, and overlaps. \label{tab:important}}
{
\noindent{\footnotesize{* YOLOv11n = YOLOv11-nano; co-detr = co\_dino\_5scale\_r50\_8xb2\_1x; faster-rcnn = faster-rcnn\_x101-64x4d\_fpn\_ms-3x}} \\
}
\end{center}
\end{table}

\section{Statistical Analysis of Patching Method}
\label{appendix:statisticalanalysis}

This study employs regression models as a means of modelling the collected data. By doing so, relationships concerning the hyperparameter choices for the patching scheme and model selection can be revealed, which this study may then attempt to explain. It also provides the opportunity to extrapolate to threshold-overlap value pairs not explicitly tested. 

To fit a model to the data in Table~\ref{tab:important}, fitting a linear regression model is a conceptually simple starting point. It expresses the target variable, such as mAP@IoU=0.5 or mAP@IoU=0.5:0.95, as a linear combination of predictor variables (and which may include terms that are functions of variables). These input variables may be categorical factors, such as the model, or continuous numerical values such as threshold or overlap. An additive model (Eq.~\ref{eq:additive_05}) can be fit, in which the target variable is expressed purely as a linear combination of input variables. However, this means that the model must use a constant slope for each variable, which often fails to capture correlations, behaviours that change based on the value of another variable. For these, interaction models are introduced (Eqs.~\ref{eq:interaction_05},~\ref{eq:interaction_full_05}) with interaction terms that are a product of multiple variables. 

\begin{figure}[!htbp]
    \centering
    \includegraphics[width=0.65\linewidth]{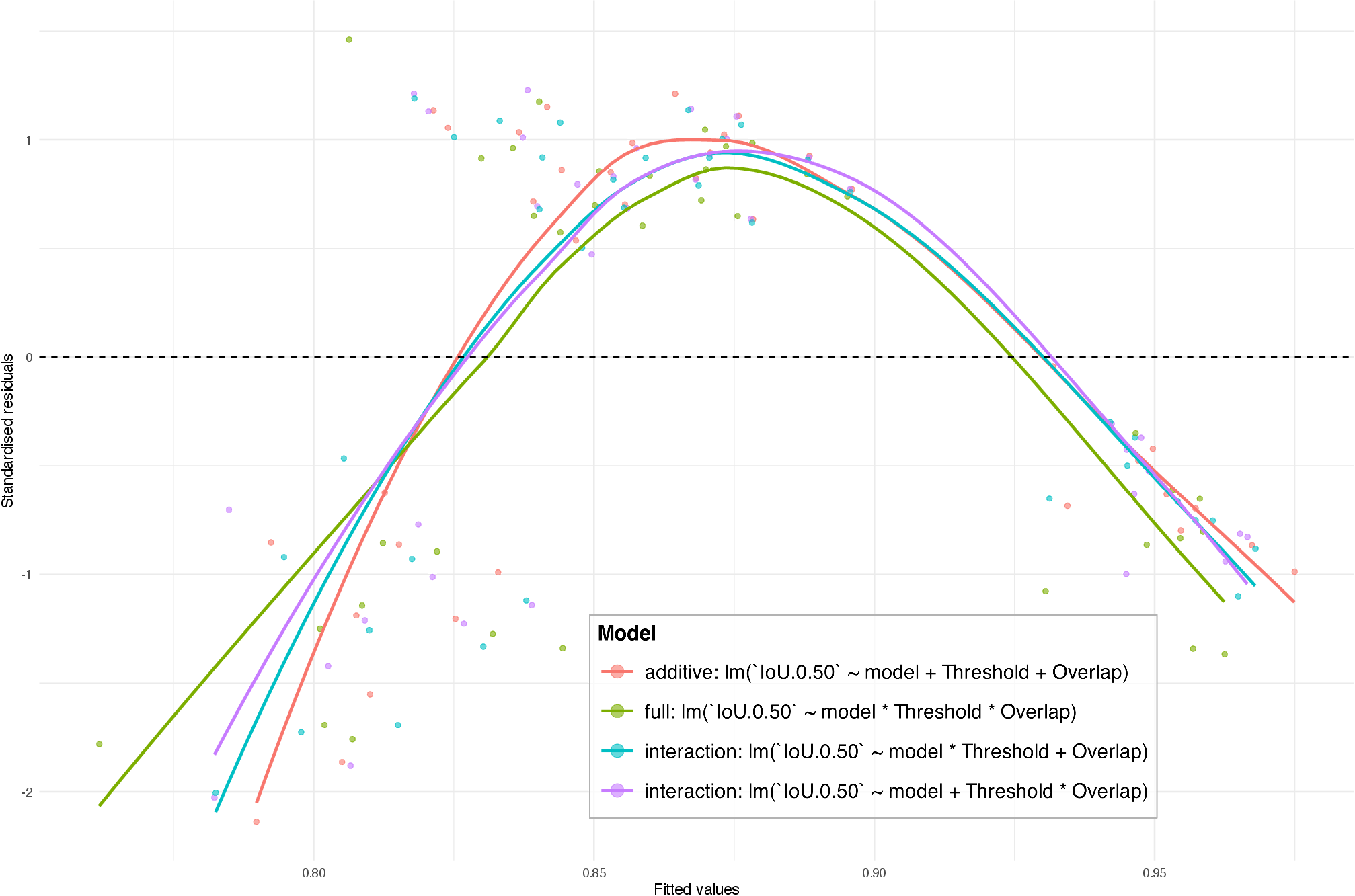}
    \caption{Residuals vs. Fitted plot of the four basic linear models, including the curves of best fit between the residual/fitted points. The curvature indicates a non-linear relationship between the output variable (mAP) and one of the predictor variables.}
    \label{fig:mlr_threemodels_goodcopy_IoU05_lm_residualsvsfitted.eps}
\end{figure}

One diagnostic plot to determine goodness of fit is the residuals vs. fitted plot (Fig.~\ref{fig:mlr_threemodels_goodcopy_IoU05_lm_residualsvsfitted.eps}), where the vertical axis represents the error between the mAP of existing data points and the mAP that the fitted model predicts based on those points. In Fig.~\ref{fig:mlr_threemodels_goodcopy_IoU05_lm_residualsvsfitted.eps}, it appears that the line of best fit of the residuals for all linear models is a parabola, and that all data points appear to fall into three clusters. This is a clear indication that mAP is non-linear in relation to at least one of the input variables. It was also noted in the data (Table~\ref{tab:important}) that the residual was negative if and only if the threshold is 0.1 or 1.0 (the extreme values of threshold), an indication that threshold may have a higher-degree relationship.

To compensate for this, the threshold variable can be represented as a polynomial (Eq.~\ref{eq:additive_poly_05},\ref{eq:interaction_poly_05}); several terms of varying degrees all have their coefficients fit. Alternatively, a spline (Eq.~\ref{eq:spline_05},\ref{eq:spline_full_05}) is a piecewise linear that can, given a predefined inflection point (a "knot"), each linear piece can have its own slope. 

\begin{figure}[!htbp]
    \centering
    \includegraphics[width=0.9\linewidth]{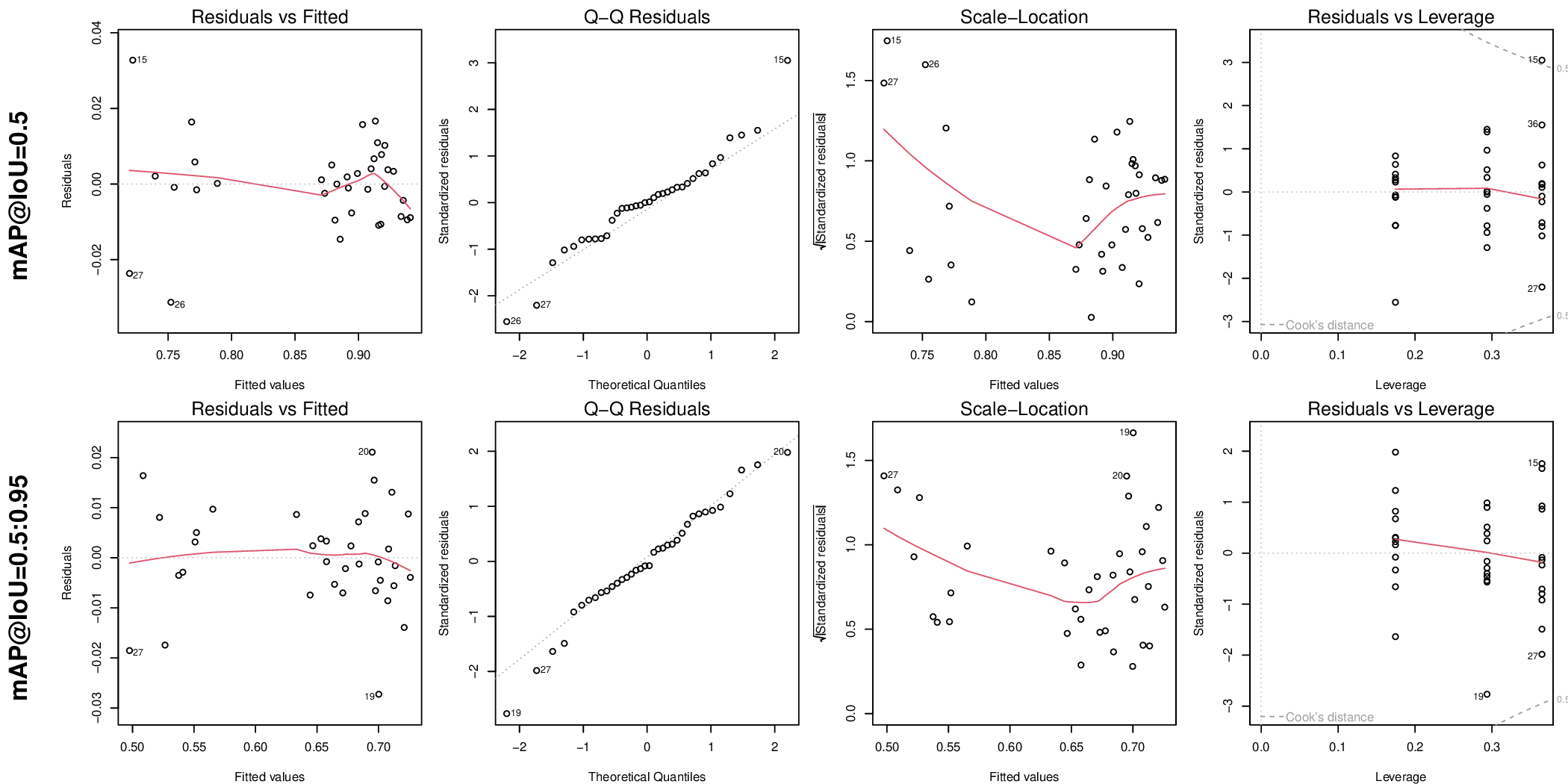}
    \caption{Diagnostic plots for Eq.~\ref{eq:interaction_poly_05} (top row) and Eq.~\ref{eq:interaction_poly_0595}.}
    \label{fig:mlr_threemodels_goodcopy_interaction_poly_diagnostics.eps}
\end{figure}

As determined empirically, letting the threshold be represented by a degree-2 polynomial was insufficient at fixing the residuals vs. fitted plot, but letting it be a  degree-3 polynomial (a cubic) provided a good result, as shown in Fig.~\ref{fig:mlr_threemodels_goodcopy_interaction_poly_diagnostics.eps}. This figure used the same model as those in Figs.~\ref{fig:linreg50},\ref{fig:linreg95}; these are, respectively, Eqs.~\ref{eq:interaction_poly_05},\ref{eq:interaction_poly_0595}. Cubic representation may be appropriate since it can handle the flat sections of mAP at the larger threshold values (Figs.~\ref{fig:barchart50},\ref{fig:barchart95}). The difference in behaviour between Threshold = 0.1 and higher threshold values in Figs.~\ref{fig:linreg50},\ref{fig:linreg95} inspired the decision to set the spline knot to 0.15. 

\begin{figure}[!htbp]
    \centering
    \includegraphics[width=0.9\linewidth]{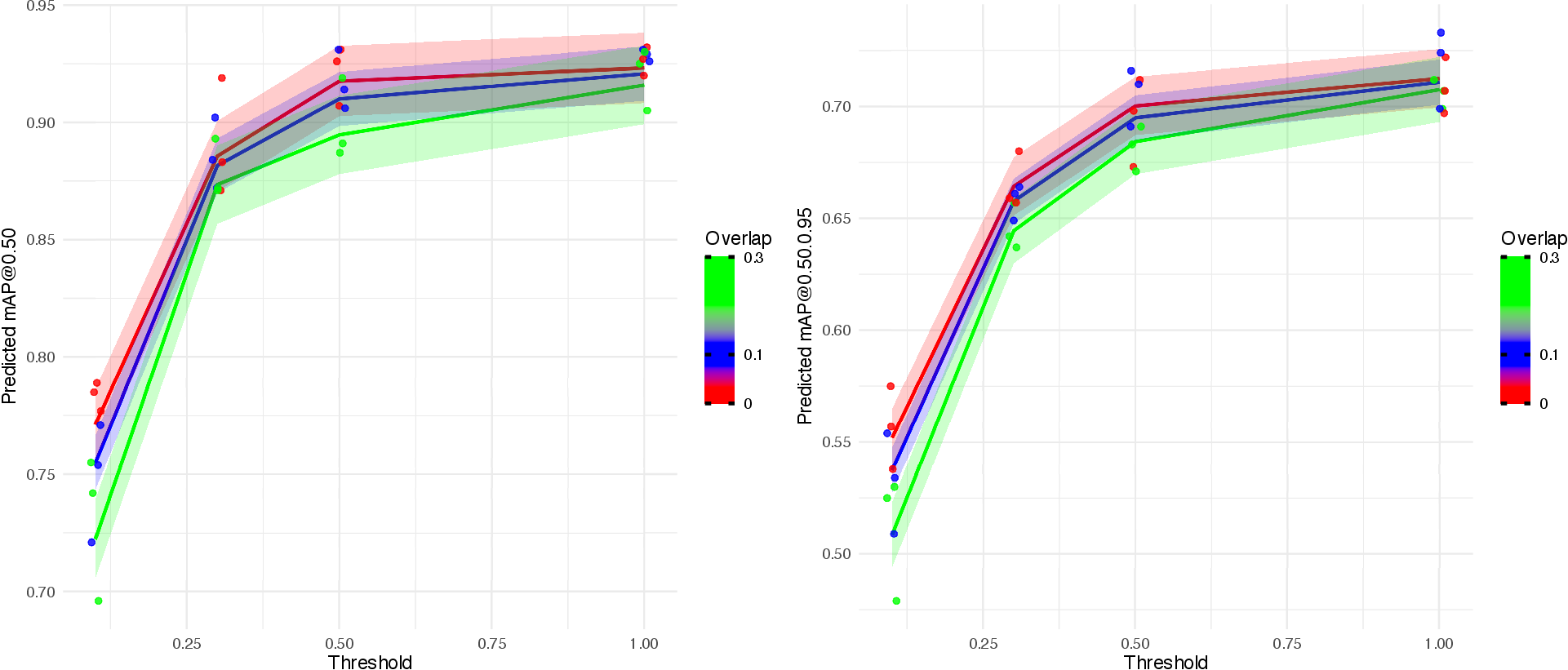}
    \caption{For each Overlap value (0.0, 0.1, 0.3), a plot of mAP against various threshold values for Eqs.~\ref{eq:interaction_poly_05},\ref{eq:interaction_poly_0595}. The dots are real data points from Table~\ref{tab:important}, the lines indicate the function that the models use to fit the points (note the non-linearity, and that they differ somewhat by overlap value), and the shaded regions around the line indicate 95\% CI from the RMSE (Table~\ref{tab:threemodels_modelselection}).}
    \label{fig:mlr_threemodels_goodcopy_interaction_poly_predictions.eps}
\end{figure}

To select the best model, a variety of experiments were run (Table~\ref{tab:threemodels_modelselection}). The clear winner is the interaction model with the polynomial threshold term, Eq.~\ref{eq:interaction_poly_05}. Its Adj. $R^2$ indicated that about 96\% of the variance in the output variable was predicted by this model's inputs, while taking number of parameters into consideration, the highest among the models. CV RMSE, which measures the model's ability to fit unseen data and is therefore a highly important measure, also favoured this model. The diagnostic plots for this polynomial interaction model (Fig.~\ref{fig:mlr_threemodels_goodcopy_interaction_poly_diagnostics.eps}) showcase good behaviour for both IoU=0.5 and IoU=0.5:0.95; for predicting each point, this model's errors are low and the model was not being pulled by outliers. Throughout the Fig.~\ref{fig:mlr_threemodels_goodcopy_interaction_poly_diagnostics.eps} plots, the residuals remain close to the theoretical residuals (or 0, if normalized), indicating that the data collection was sound and the model fits the data well. In Fig.~\ref{fig:mlr_threemodels_goodcopy_interaction_poly_predictions.eps}, the polynomial interaction model's threshold curve was plotted against the data at various overlap values, clearly exhibiting the curvature which a non-linear term was required to model. The candidates are as follows:

\begin{equation}
    \verb|model_additive = lm(IoU.0.50 ~ model + Threshold + Overlap)|
    \label{eq:additive_05}
\end{equation}

\begin{equation}
    \verb|model_additive_poly = lm(mAP@IoU=0.5 ~ model + poly(Threshold, 3) + Overlap)|
    \label{eq:additive_poly_05}
\end{equation}

\begin{equation}
    \verb|model_interaction = lm(IoU.0.50 ~ model + Threshold * Overlap)|
    \label{eq:interaction_05}
\end{equation}

\begin{equation}
    \verb|model_interaction_full = lm(IoU.0.50 ~ model * Threshold * Overlap)|
    \label{eq:interaction_full_05}
\end{equation}

\begin{equation}
    \verb|model_interaction_poly = lm(mAP@IoU=0.5 ~ model + poly(Threshold, 3) * Overlap)|
    \label{eq:interaction_poly_05}
\end{equation}

\begin{equation}
    \verb|model_spline = lm(IoU.0.50 ~ model + ns(Threshold, knots = c(0.15)) * Overlap)|
    \label{eq:spline_05}
\end{equation}

\begin{equation}
    \verb|model_spline_full <- lm(IoU.0.50 ~ model + ns(Threshold, knots = c(0.15)) * Overlap)|
    \label{eq:spline_full_05}
\end{equation}

\begin{equation}
    \verb|additive = lm(mAP@IoU=0.5:0.95 ~ model + poly(Threshold, 3) + Overlap)|
    \label{eq:additive_poly_0595}
\end{equation}

\begin{equation}
    \verb|interaction = lm(mAP@IoU=0.5:0.95 ~ model + poly(Threshold, 3) * Overlap)|
    \label{eq:interaction_poly_0595}
\end{equation}

A few analyses can be done to confirm that the model is not overfitting, a potential consequence of using three input variables and their interactions to model the only 36 points in Table~\ref{tab:important}. Based on ANOVA, the p-value is about 0.05, on the border between the smaller additive model from being sufficient and the interactive model being needed (Table~\ref{tab:threemodels_modelselection}). This study uses the interaction model since an in-depth view of the threshold-overlap relationship is useful for analyzing the patching method. To ensure that the model does not overfit, this study also conducted an RMSE simulation. For both the additive model Eq.~\ref{eq:additive_poly_05} and interaction model Eq.~\ref{eq:interaction_poly_05}, a 10-fold cross validation was performed. Since the RMSE of the interaction was lower, this suggests better fit. In addition, this also produced 95\% CIs for the interaction models in Eqs.~\ref{eq:interaction_poly_05},\ref{eq:interaction_poly_0595}, displayed in Fig.~\ref{fig:mlr_threemodels_goodcopy_interaction_poly_predictions.eps}. Since four distinct threshold values were measured, and it is a well-known result that for any $n$ points a $n-1$-degree polynomial is sufficient to pass through all points exactly, Fig.~\ref{fig:mlr_threemodels_goodcopy_interaction_poly_predictions.eps} helps to show that while the model fits well, it is not overfitting.

\begingroup                 
\setlength{\tabcolsep}{3pt} 

\begin{table}[htbp]
\centering
\begin{tabular}{@{}l l c c c c c c@{}}
\hline
Model & Formula* & df & Adj.\ $R^{2}$ & CV RMSE & Residual SD & AIC & BIC \\
\hline
model\_additive [\ref{eq:additive_05}] & M $+$ T $+$ O                             &  \textbf{6} & 0.545 & 0.0516 & 0.0481 & $-109.68$ & $-100.18$ \\
\hline
model\_additive\_poly [\ref{eq:additive_poly_05}] & M $+$ poly(T, 3) $+$ O                    &  8 & 0.957 & 0.0148 & 0.0148 & $-193.06$ & $\mathbf{-180.40}$ \\
\hline
model\_interaction [\ref{eq:interaction_05}] & M $+$ T $\times$ O                        &  7 & 0.536 & 0.0506 & 0.0486 & $-108.14$ & $ -97.05$ \\
\hline
model\_interaction\_full [\ref{eq:interaction_full_05}] & M $\times$ T $\times$ O                   & 13 & 0.437 & 0.0689 & 0.0535 & $ -97.19$ & $ -76.61$ \\
\hline
\textbf{model\_interaction\_poly [\ref{eq:interaction_poly_05}]} & \textbf{M $+$ poly(T, 3) $\times$ O} & 11 & \textbf{0.964} & \textbf{0.0141} & \textbf{0.0135} & $\mathbf{-197.64}$ & -180.22 \\
\hline
model\_spline [\ref{eq:spline_05}] & M $+$ ns(T, knot = 0.15) $\times$ O       &  9 & 0.937 & 0.0194 & 0.0179 & $-178.65$ & $-164.39$ \\
\hline
model\_spline\_full [\ref{eq:spline_full_05}] & M $\times$ ns(T, knot = 0.15) $\times$ O  & 19 & 0.935 & 0.0257 & 0.0182 & $-173.37$ & $-143.28$ \\
\hline
\end{tabular}
\vspace{8pt}
\caption{Model selection metrics.  Lower is better for RMSE, AIC, and BIC; higher is better for $\text{Adj.\ }R^{2}$.}
{
\noindent{\footnotesize{* M = Model, T = Threshold, O = Overlap}} \\
}
\label{tab:threemodels_modelselection}
\end{table}

\newpage

\section{YOLOv11 Data}
\label{appendix:YOLOv11DataTable}

\begingroup                 
\setlength{\tabcolsep}{2pt} 
\renewcommand{\arraystretch}{1.0} 

\begin{table}[!htbp]
\centering
\begin{tabular}{|c|c|c|c|c|c|c|c|c|c|c|c|c|c|}
\hline
\multicolumn{3}{|c|}{~} & \multicolumn{2}{c|}{Box} & \multicolumn{2}{c|}{mAP@IoU} & \multicolumn{3}{c|}{Speed} & \multicolumn{2}{c|}{Training} & \multicolumn{2}{c|}{Validation}\\
\hline
\textbf{Threshold} & \textbf{Overlap} & \textbf{Size} & \textbf{P} & \textbf{R} & \textbf{50} & \textbf{50:95} & \textbf{Preproc} & \textbf{Inference} & \textbf{Postproc} & \textbf{Images} & \textbf{Boxes} & \textbf{Images} & \textbf{Boxes} \\
\hline
0.1 & 0.0 & large & 0.875 & 0.637 & 0.717 & 0.513 & 0.9 & 1.4 & 4.3 & 2020 & 3578 & 158 & 306 \\ \hline
0.1 & 0.0 & medium & 0.875 & 0.688 & 0.753 & 0.542 & 0.6 & 1.1 & 2.0 & 2020 & 3578 & 160 & 276 \\ \hline
0.1 & 0.0 & small & 0.929 & 0.803 & 0.873 & 0.655 & 0.7 & 0.8 & 2.6 & 2020 & 3578 & 192 & 351 \\ \hline
0.1 & 0.1 & large & 0.879 & 0.636 & 0.724 & 0.523 & 0.7 & 1.5 & 4.1 & 2275 & 4197 & 199 & 368 \\ \hline
0.1 & 0.1 & medium & 0.848 & 0.677 & 0.753 & 0.561 & 0.5 & 1.2 & 2.6 & 2275 & 4197 & 188 & 325 \\ \hline
0.1 & 0.1 & small & 0.877 & 0.782 & 0.861 & 0.671 & 0.6 & 0.9 & 2.7 & 2275 & 4197 & 219 & 411 \\ \hline
0.1 & 0.3 & large & 0.898 & 0.643 & 0.743 & 0.547 & 0.7 & 1.5 & 4.0 & 2975 & 6244 & 253 & 440 \\ \hline
0.1 & 0.3 & medium & 0.864 & 0.694 & 0.768 & 0.578 & 0.5 & 1.2 & 2.5 & 2975 & 6244 & 238 & 409 \\ \hline
0.1 & 0.3 & small & 0.893 & 0.794 & 0.867 & 0.689 & 0.6 & 0.9 & 2.7 & 2975 & 6244 & 292 & 566 \\ \hline
0.2 & 0.0 & large & 0.870 & 0.601 & 0.690 & 0.485 & 0.9 & 1.5 & 4.2 & 2020 & 3578 & 158 & 306 \\ \hline
0.2 & 0.0 & medium & 0.860 & 0.646 & 0.720 & 0.511 & 0.6 & 1.2 & 2.0 & 2020 & 3578 & 160 & 276 \\ \hline
0.2 & 0.0 & small & 0.908 & 0.781 & 0.850 & 0.629 & 0.7 & 0.8 & 2.6 & 2020 & 3578 & 192 & 351 \\ \hline
0.2 & 0.1 & large & 0.880 & 0.640 & 0.730 & 0.528 & 0.7 & 1.5 & 4.1 & 2275 & 4197 & 199 & 368 \\ \hline
0.2 & 0.1 & medium & 0.851 & 0.669 & 0.747 & 0.556 & 0.5 & 1.2 & 2.6 & 2275 & 4197 & 188 & 325 \\ \hline
0.2 & 0.1 & small & 0.880 & 0.779 & 0.861 & 0.671 & 0.6 & 0.9 & 2.7 & 2275 & 4197 & 219 & 411 \\ \hline
0.2 & 0.3 & large & 0.897 & 0.640 & 0.738 & 0.542 & 0.7 & 1.5 & 4.0 & 2975 & 6244 & 253 & 440 \\ \hline
0.2 & 0.3 & medium & 0.859 & 0.685 & 0.761 & 0.570 & 0.5 & 1.2 & 2.5 & 2975 & 6244 & 238 & 409 \\ \hline
0.2 & 0.3 & small & 0.889 & 0.792 & 0.862 & 0.686 & 0.6 & 0.9 & 2.7 & 2975 & 6244 & 292 & 566 \\ \hline
0.3 & 0.0 & large & 0.858 & 0.552 & 0.648 & 0.457 & 0.9 & 1.5 & 4.2 & 2020 & 3578 & 158 & 306 \\ \hline
0.3 & 0.0 & medium & 0.846 & 0.598 & 0.689 & 0.486 & 0.6 & 1.2 & 2.0 & 2020 & 3578 & 160 & 276 \\ \hline
0.3 & 0.0 & small & 0.892 & 0.759 & 0.833 & 0.612 & 0.7 & 0.8 & 2.6 & 2020 & 3578 & 192 & 351 \\ \hline
0.3 & 0.1 & large & 0.868 & 0.583 & 0.684 & 0.490 & 0.7 & 1.5 & 4.1 & 2275 & 4197 & 199 & 368 \\ \hline
0.3 & 0.1 & medium & 0.840 & 0.626 & 0.713 & 0.521 & 0.5 & 1.2 & 2.6 & 2275 & 4197 & 188 & 325 \\ \hline
0.3 & 0.1 & small & 0.873 & 0.774 & 0.856 & 0.666 & 0.6 & 0.9 & 2.7 & 2275 & 4197 & 219 & 411 \\ \hline
0.3 & 0.3 & large & 0.889 & 0.611 & 0.715 & 0.524 & 0.7 & 1.5 & 4.0 & 2975 & 6244 & 253 & 440 \\ \hline
0.3 & 0.3 & medium & 0.852 & 0.663 & 0.743 & 0.553 & 0.5 & 1.2 & 2.5 & 2975 & 6244 & 238 & 409 \\ \hline
0.3 & 0.3 & small & 0.881 & 0.786 & 0.860 & 0.684 & 0.6 & 0.9 & 2.7 & 2975 & 6244 & 292 & 566 \\ \hline
0.4 & 0.0 & large & 0.843 & 0.497 & 0.602 & 0.428 & 1.0 & 1.6 & 4.3 & 2020 & 3578 & 158 & 306 \\ \hline
0.4 & 0.0 & medium & 0.829 & 0.545 & 0.646 & 0.456 & 0.7 & 1.3 & 2.1 & 2020 & 3578 & 160 & 276 \\ \hline
0.4 & 0.0 & small & 0.876 & 0.723 & 0.803 & 0.589 & 0.7 & 0.9 & 2.6 & 2020 & 3578 & 192 & 351 \\ \hline
0.4 & 0.1 & large & 0.856 & 0.530 & 0.633 & 0.450 & 0.7 & 1.6 & 4.1 & 2275 & 4197 & 199 & 368 \\ \hline
0.4 & 0.1 & medium & 0.829 & 0.582 & 0.669 & 0.477 & 0.5 & 1.3 & 2.7 & 2275 & 4197 & 188 & 325 \\ \hline
0.4 & 0.1 & small & 0.864 & 0.746 & 0.833 & 0.648 & 0.6 & 0.9 & 2.8 & 2275 & 4197 & 219 & 411 \\ \hline
0.4 & 0.3 & large & 0.879 & 0.573 & 0.689 & 0.503 & 0.7 & 1.6 & 4.2 & 2975 & 6244 & 253 & 440 \\ \hline
0.4 & 0.3 & medium & 0.835 & 0.629 & 0.712 & 0.521 & 0.6 & 1.3 & 2.5 & 2975 & 6244 & 238 & 409 \\ \hline
0.4 & 0.3 & small & 0.868 & 0.780 & 0.860 & 0.687 & 0.6 & 0.9 & 2.9 & 2975 & 6244 & 292 & 566 \\ \hline
0.5 & 0.0 & large & 0.818 & 0.441 & 0.553 & 0.390 & 1.0 & 1.6 & 4.3 & 2020 & 3578 & 158 & 306 \\ \hline
0.5 & 0.0 & medium & 0.801 & 0.485 & 0.595 & 0.418 & 0.7 & 1.3 & 2.1 & 2020 & 3578 & 160 & 276 \\ \hline
0.5 & 0.0 & small & 0.854 & 0.688 & 0.773 & 0.568 & 0.8 & 0.9 & 2.6 & 2020 & 3578 & 192 & 351 \\ \hline
0.5 & 0.1 & large & 0.832 & 0.492 & 0.590 & 0.417 & 0.7 & 1.6 & 4.2 & 2275 & 4197 & 199 & 368 \\ \hline
0.5 & 0.1 & medium & 0.803 & 0.526 & 0.620 & 0.436 & 0.6 & 1.3 & 2.7 & 2275 & 4197 & 188 & 325 \\ \hline
0.5 & 0.1 & small & 0.854 & 0.726 & 0.806 & 0.606 & 0.6 & 0.9 & 2.8 & 2275 & 4197 & 219 & 411 \\ \hline
0.5 & 0.3 & large & 0.857 & 0.540 & 0.645 & 0.462 & 0.7 & 1.6 & 4.1 & 2975 & 6244 & 253 & 440 \\ \hline
0.5 & 0.3 & medium & 0.815 & 0.593 & 0.682 & 0.492 & 0.6 & 1.3 & 2.5 & 2975 & 6244 & 238 & 409 \\ \hline
0.5 & 0.3 & small & 0.859 & 0.785 & 0.860 & 0.689 & 0.6 & 0.9 & 2.9 & 2975 & 6244 & 292 & 566 \\ \hline
\end{tabular}
\vspace{8pt}
\caption{YOLOv11 experimental results used in Appendix~\ref{appendix:YOLOpatchsize}. For each row, YOLOv11 was trained on the dataset version with hyperparameter values \textbf{Threshold} and \textbf{Overlap}, whose training set has Training:\textbf{Images} patches with a total of Training:\textbf{Boxes} annotations. This training set includes patches of all three sizes. Each row's validation set was the subset of the validation set where the patches are of size \textbf{Size}.}
\label{tab:yolo_patch_size}
\end{table}
\endgroup

\section{Analysis of Patch Size on YOLOv11}
\label{appendix:YOLOpatchsize}

Having established the superiority of YOLOv11 \cite{Jocher_Ultralytics_YOLO_2023} to the other models in this context, this section investigates the effect of potential confounding variables to assert that varying the patching method's hyperparameters explains much more of the variance than other variables or random chance. The focus is on the performance of YOLOv11 on patches of various scales, or "sizes", once it has been trained on all sizes. 

\subsection{Candidate Linear Models}

\begin{equation}
    \verb|model_additive <- lm(mAP50 ~ size + threshold + overlap + Images)|
    \label{eq:yolo_additive}
\end{equation}

\begin{equation}
    \verb|model_interaction <- lm(mAP50 ~ size * threshold * overlap + Images)|
    \label{eq:yolo_interaction}
\end{equation}

\begin{equation}
    \verb|model_interaction_poly <- lm(mAP50 ~ size * poly(threshold, 3) * overlap + Images)|
    \label{eq:yolo_interaction_poly}
\end{equation}

\begin{equation}
    \verb|model_spline <- lm(mAP50 ~ size * ns(threshold, knots = c(0.15)) * overlap + Images)|
    \label{eq:yolo_spline}
\end{equation}

\subsection{Images vs. Instances}

One basic heuristic in deep learning is that more training data typically leads to better results, since the models see more examples from which to create a more accurate loss surface. This section seeks to assert that changes in understanding are not merely caused by a larger quantity of data.

The number of images the model was trained on and the number of annotations (after filtering according to the Threshold) was considered. Of note is the fact that, while models evaluated on the small, medium and large images were all trained using all sizes, whether the validation images or the training images were used in the linear models makes little difference. This is because the image count has a very minor effect on the predictions, a much weaker effect than the threshold and overlap hyperparameters.

To determine which variables to keep, cross-validated feature pruning using Recursive Feature Elimination (RFE) was performed on the models (Eqs.~\ref{eq:yolo_additive},\ref{eq:yolo_interaction},\ref{eq:yolo_interaction_poly},\ref{eq:yolo_spline}). On all models, Images was more statistically significant than Instances, with Images very slightly negatively correlated with mAP and Instances very slightly positively correlated with mAP. Although patches without instances were filtered, this finding represents the shortage of positive areas; the vast majority of each patch is still background. However, image count cannot be completely discounted, and this very slight but significant negative correlation between training image count and mAP may be due to the large negative areas (Fig.~\ref{fig:spline})

\subsection{Statistical Model Effectiveness}

To determine the best model, several experiments were run and are shown in Table~\ref{table:yolodiagnostics}. To make the decision, several priors are worth considering. For one, with only $n = 36$ data points, and each model using $36 - \textbf{Res.df}$ parameters as a result of the different variables needing incorporation, overfitting is the critical issue and metrics that check for fit on new data (\textbf{CV RMSE}) must be prioritized above the others. Additionally, it was established in Appendix~\ref{appendix:statisticalanalysis} that an interaction model with a higher-degree threshold term provided better results than the more basic models.  

When considering the model statistics in Table~\ref{table:yolodiagnostics}, the Additive model's adjusted R-squared ($\textbf{Adj.~}\mathbf{R^2}$) suggests that about 92\% of the variance is explained by the variables, so it may be slightly underfitting. The Interaction model wins in terms of $F$~\textbf{vs. prev.} and $p$\textbf{-value} but had the worst \textbf{CV RMSE}. This is driven by Interaction's per-parameter usefulness, but is captured by noise while not modelling nonlinear relationships. The Poly model showed stellar performance on in-sample statistics, though poor \textbf{CV RMSE} (and fitting 25 parameters on just 36 points) demonstrates overfitting. Spline was respectable overall, and its \textbf{CV RMSE} shows great ability to generalize. Though Figs.~\ref{fig:barchart50},\ref{fig:barchart95} illustrate that the three models perform similarly, Table~\ref{tab:important} may be more complex than Table~\ref{tab:yolo_patch_size} since it included more models and hence the Spline was sufficient.

\begingroup                 
\setlength{\tabcolsep}{3pt} 
\renewcommand{\arraystretch}{1.1} 
\begin{table}[ht]
\centering
\begin{tabular}{@{}|c|c|c|c|c|c|c|c|c|c|@{}} 
\hline
~ & \multicolumn{2}{c|}{Model} & \multicolumn{7}{c|}{ANOVA} \\
\hline
\textbf{Model} & $\textbf{Adj.~}\mathbf{R^2}$ & $p$\textbf{-value} & \textbf{Res.\,df} & \textbf{RSS} & \textbf{CV RMSE} & \textbf{AIC} & \textbf{BIC} & $F$~\textbf{vs.\ prev.} & \textbf{$p$-value} \\
\hline
Additive [\ref{eq:yolo_additive}]   & 0.9215 & $\mathbf{<\!2.0\text{\textbf{e}}{-16}}$ & 30 & 0.014912 & 0.0215 & -164.25 & -153.16 & --   & --        \\ \hline
Interaction [\ref{eq:yolo_interaction}] & 0.9448 & 1.20e-13 & 23 & 0.008039 & 0.0259 & -172.49 & -150.32 & \textbf{6.27} & \textbf{3.84e-03} \\ \hline
Poly [\ref{eq:yolo_interaction_poly}]        & \textbf{0.9753} & 1.74e-08 & 11 & \textbf{0.001723} & 0.0266 & \textbf{-203.95} & \textbf{-162.77} & 3.36 & 2.68e-02 \\ \hline
\textbf{Spline [\ref{eq:yolo_spline}]}     & 0.9588 & 5.56e-11 & 17 & 0.004430 & \textbf{0.0190} & -181.94 & -150.27 & 2.88 & 6.12e-02 \\ \hline
\end{tabular}
\vspace{8pt}
\caption{Comparison of candidate models for predicting mAP@IoU=0.5.  Lower values are better for RSS, CV RMSE, AIC, and BIC; higher values are better for the $F$ statistic. Bold entries indicate the best value in each column. \label{table:yolodiagnostics}}
\end{table}
\endgroup

In Fig.~\ref{fig:spline}, \verb|sizemedium| and \verb|sizesmall| are significant and positively correlated with mAP. This showcases the degree to which zoomed-in patches make the detection task easier. Compared to large and medium patches, performance on the small patches is about 14\% higher. However, on some threshold-overlap pairs, even the performance on the large patches is very high. The proportion of the dataset that should be each of small, medium and large during the training phase is unexplored by this work, but this paper shows that even if the onboard algorithm only passes large patches to the detector (for speed or memory reasons), very strong performance can be attained even when as little as 30\% of a moose is visible.

\begin{figure}
    \centering
    \tiny
    \begin{verbatim}
                        Coefficients:
                                                                             Estimate Std. Error t value Pr(>|t|)    
                        (Intercept)                                         1.008e+00  7.267e-02  13.872 1.06e-10 ***
                        sizemedium                                          4.667e-02  1.867e-02   2.500 0.022967 *  
                        sizesmall                                           1.373e-01  1.867e-02   7.356 1.12e-06 ***
                        ns(threshold, knots = c(0.15))1                     1.374e-01  6.301e-02   2.181 0.043511 *  
                        ns(threshold, knots = c(0.15))2                     3.168e-02  2.637e-02   1.201 0.246047    
                        overlap                                             4.580e-01  1.362e-01   3.361 0.003704 ** 
                        count_images_train                                 -1.455e-04  3.634e-05  -4.004 0.000918 ***
                        sizemedium:ns(threshold, knots = c(0.15))1          4.655e-03  4.928e-02   0.094 0.925852    
                        sizesmall:ns(threshold, knots = c(0.15))1          -1.356e-01  4.928e-02  -2.752 0.013597 *  
                        sizemedium:ns(threshold, knots = c(0.15))2         -1.471e-02  2.411e-02  -0.610 0.549927    
                        sizesmall:ns(threshold, knots = c(0.15))2          -7.502e-02  2.411e-02  -3.112 0.006345 ** 
                        sizemedium:overlap                                 -1.004e-01  1.023e-01  -0.982 0.339914    
                        sizesmall:overlap                                  -2.320e-01  1.023e-01  -2.269 0.036566 *  
                        ns(threshold, knots = c(0.15))1:overlap            -5.499e-01  2.182e-01  -2.520 0.022035 *  
                        ns(threshold, knots = c(0.15))2:overlap            -1.770e-01  9.940e-02  -1.781 0.092788 .  
                        sizemedium:ns(threshold, knots = c(0.15))1:overlap  2.905e-01  2.699e-01   1.076 0.296875    
                        sizesmall:ns(threshold, knots = c(0.15))1:overlap   5.626e-01  2.699e-01   2.084 0.052539 .  
                        sizemedium:ns(threshold, knots = c(0.15))2:overlap  9.355e-02  1.321e-01   0.708 0.488277    
                        sizesmall:ns(threshold, knots = c(0.15))2:overlap   2.137e-01  1.321e-01   1.618 0.124094    
                        ---
                        Signif. codes:  0 ‘***’ 0.001 ‘**’ 0.01 ‘*’ 0.05 ‘.’ 0.1 ‘ ’ 1
                        
                        Residual standard error: 0.01614 on 17 degrees of freedom
                        Multiple R-squared:   0.98,	Adjusted R-squared:  0.9588 
                        F-statistic:  46.3 on 18 and 17 DF,  p-value: 5.556e-11
    \end{verbatim}
    \caption{Linear fit summary, mAP@IoU=0.50 of the Spline model (Eq.~\ref{eq:yolo_spline}).}
    \label{fig:spline}
\end{figure}

\end{document}